\documentclass{article}
\usepackage[margin=1in]{geometry}
\usepackage{graphicx} 
\usepackage[bibstyle=numeric, citestyle=authoryear]{biblatex}
\usepackage[parfill]{parskip}
\usepackage[colorlinks, urlcolor=blue!80!black, linkcolor=blue!50!black, citecolor=red!80!black]{hyperref}
\usepackage{courier}
\usepackage{makecell}
\usepackage{xcolor}

\title{Enhancing Q\&A with Domain-Specific Fine-Tuning and Iterative Reasoning: A Comparative Study}
\author{
    Zooey Nguyen\thanks{Aitomatic, Inc. (except as noted, all authors are from Aitomatic)}
    \and Anthony Annunziata\thanks{IBM Research}
    \and Vinh Luong
    \and Sang Dinh
    \and Quynh Le
    \and Anh Hai Ha
    \and Chanh Le
    \and Hong An Phan
    \and Shruti Raghavan
    \and Christopher Nguyen
}

\date{April 2024}

\begin{document}
\maketitle

\begin{abstract}
This paper investigates the impact of domain-specific model fine-tuning and of reasoning mechanisms on the performance of question-answering (Q\&A) systems powered by large language models (LLMs) and Retrieval-Augmented Generation (RAG). Using the \texttt{FinanceBench} SEC financial filings dataset, we observe that, for RAG, combining a fine-tuned embedding model with a fine-tuned LLM achieves better accuracy than generic models, with relatively greater gains attributable to fine-tuned embedding models. Additionally, employing reasoning iterations on top of RAG delivers an even bigger jump in performance, enabling the Q\&A systems to get closer to human-expert quality. We discuss the implications of such findings, propose a structured technical design space capturing major technical components of Q\&A AI, and provide recommendations for making high-impact technical choices for such components. We plan to follow up on this work with actionable guides for AI teams and further investigations into the impact of domain-specific augmentation in RAG and into agentic AI capabilities such as advanced planning and reasoning.\footnote{This collaborative work is facilitated by the AI Alliance \url{https://thealliance.ai}, in the interest of open development, open science and open AI. All of the authors are affiliated with the AI Alliance. Except for specifics disclosed in this paper, all intellectual properties of the work belong to respective owners.}
\end{abstract}


\section{Introduction}
AI-powered question-answering (Q\&A) systems have emerged as important tools, alongside established search technologies, to enable quick access to relevant information and knowledge from large digital sources that are complex and time-consuming for humans to navigate. Advancements in large language models (LLMs) have revolutionized the field of Q\&A, with models like GPT-3 \parencite{brown2020language}, BERT \parencite{devlin2019bert}, and RoBERTa \parencite{liu2019roberta} demonstrating remarkable abilities in understanding and generating human-like text. However, the effectiveness of such models in handling domain-specific questions that require specialized knowledge is limited.

Retrieval-augmented generation (RAG) techniques, which combine information retrieval and generative models \parencite{lewis2021retrievalaugmented}, have shown promise in boosting the quality of LLM output in Q\&A tasks. RAG systems leverage the strengths of both retrieval and generation components to provide contextually relevant and informative responses. While there is a lack of established quantification of RAG accuracy, early findings suggest that generic RAG does not perform well in complex domains such as finance. In one instance, RAG based on generic LLMs such as GPT-4-Turbo fails to answer 81\% of the questions derived from Securities and Exchange Commission (SEC) financial filings \parencite{islam2023financebench}.

The underperformance of generic LLMs and RAG in domain-specific Q\&A has motivated us to research into and create methods to adapt and extend such models and techniques. In this paper, we describe and quantify the gains in accuracy from two major methods: model fine-tuning and iterative reasoning.

Fine-tuning is a way to adapt language models to specific domains and tasks \parencites{devlin2019bert, liu2019roberta} by training them on domain-specific data and having them capture the nuances and intricacies of a particular field. In a typical RAG workflow, there are two principal models that can be considered for fine-tuning: the Embedding Model, whose tasks are indexing the information in the corpus and retrieving information relevant to the posed question, and the Generative Model, whose task is synthesizing an answer. We have performed experiments and analyses comparing baseline generic-LLM-based RAG and RAG with one or both of fine-tuned retrieval and fine-tuned generation on the \texttt{FinanceBench} dataset, and observed promising gains from fine-tuning.

In addition, incorporating iterative reasoning and planning is crucial for intelligent agents to effectively tackle real-world, complex tasks as highlighted by \cite{lecun22path} and \cite{zhou2024selfdiscover}. Motivated by those works and by our first-hand experience in industrial applications, we have explored integrating the Observe-Orient-Decide-Act (OODA) loop \parencite{boyd86patterns}, a well-established iterative reasoning mechanism, with RAG-based Q\&A. That too has shown promising improvements, especially on highly complex questions.

By investigating the above areas, our main contributions are:

\begin{enumerate}
    \item We identify clear steps in LLM-based Q\&A workflows where further innovation can make large improvements to those workflows’ ultimate accuracy;
    \item We chart out the beginnings of a structured technical design space for the primary technical choices for each of such major steps; and
    \item We quantify the accuracy impacts of certain technical choices, with firm grounding on a concrete and sufficiently complex real-world dataset, and highlight technical configurations from which we observe large gains. Such quantification could serve developers and managers of AI systems in making informed system-design decisions for quicker success.
\end{enumerate}

The rest of the paper is organized as follows: Section \ref{background} provides an overview of related work in Q\&A AI, focusing on RAG techniques, fine-tuning strategies, and high-level planning and reasoning. Section \ref{methodology} describes our framework for improving on baseline generic RAG, proposes a structured design space for technical decision-making, and overviews the \texttt{\texttt{FinanceBench}} dataset and the technical configurations we have experimented with on that data. Section \ref{results} presents results from our experiments, followed by a discussion of our findings in Section \ref{analysis}. Finally, Section \ref{conclusion} concludes and outlines future directions.

\section{Related Work}
\label{background}

The introduction of the Transformer architecture \parencite{vaswani2017attention} laid the foundation for Q\&A AI, enabling the creation of models like BERT \parencite{devlin2019bert}, RoBERTa \parencite{liu2019roberta}, and GPT-3 \parencite{brown2020language}, now commonly known as large language models (LLMs) from circa 2021.

One of the key challenges in applying LLMs to Q\&A is handling long-form text and maintaining context over extended sequences. Techniques like Longformer \parencite{beltagy2020longformer} and Transformer-XL \parencite{dai2019transformerxl} have been developed to address this issue by introducing attention mechanisms that efficiently process longer texts. These advancements and works derived from them have paved the way for more effective retrieval and understanding of relevant information in Q\&A systems.

The seminal work by \cite{lewis2021retrievalaugmented} introduced foundational techniques of RAG, demonstrating its efficacy in knowledge-intensive NLP tasks by augmenting generative models with retrieved documents to provide contextually rich answers. Subsequent research has explored various aspects of RAG, such as the integration of retrieval and generation within LLMs \parencite{feng2023retrievalgeneration}, prompt-guided retrieval augmentation for non-knowledge-intensive tasks \parencite{guo2023promptguided}, and active retrieval augmented generation \parencite{jiang2023active}. The versatility of RAG systems has been showcased in their ability to handle diverse question types and data formats. UnifiedQA \parencite{khashabi2020unifiedqa} demonstrates how advanced RAG systems can cross format boundaries and address diverse Q\&A tasks. Datasets like HybridQA \parencite{chen2021hybridqa} have further highlighted the challenges and opportunities in dealing with hybrid data formats and emphasized the need for Q\&A systems to effectively navigate between structured and unstructured data. 

Another major R\&D area in adapting LLMs to domain-specific contexts is fine-tuning. Works like BERT \parencite{devlin2019bert} and RoBERTa \parencite{liu2019roberta} have shown the effectiveness of fine-tuning in performing tasks in certain non-generic fields. More efficient fine-tuning approaches, such as adapter layers \parencite{houlsby2019parameterefficient} and model distillation \parencite{sanh2020distilbert}, have further enhanced the practicality of domain-specific LLMs.

As the field of Q\&A continues to evolve, there is a growing emphasis on developing models that can handle complex, multi-hop reasoning. The integration of in-context learning with RAG \parencite{huang2024raven} and the exploration of unified frameworks like T5 \parencite{raffel2023exploring} represent promising directions.

Several authors have emphasized the necessity of iterative reasoning and planning in achieving real-world-capable intelligence. \cite{lecun22path} proposes a predictive world model architecture that enables abstract representation and prediction at multiple timescales, addressing uncertainty and enabling more human-like learning and reasoning. SELF-DISCOVER \parencite{zhou2024selfdiscover} is a framework that allows LLMs to self-compose task-specific reasoning structures from atomic reasoning modules, improving performance by up to 32\% on challenging reasoning benchmarks while requiring 10-40x fewer inference computations compared to inference-intensive methods. SELF-REFINE \parencite{madaan2023selfrefine}, an approach for improving initial outputs from LLMs through iterative feedback and refinement, improves accuracy by approximately 20\% on 7 diverse tasks including mathematical reasoning using GPT-3.5 and GPT-4.  To tackle the challenge of uncertainty in multi-step reasoning, Self-Evaluation Guided Beam Search for Reasoning \parencite{xie2023selfevaluation} proposes a decoding algorithm integrating self-evaluation guidance via stochastic beam search, a strategy that can pinpoint failures in logical reasoning and leads to higher consistency and robustness. 

Tree of Thoughts \parencite{yao2023tree}, a generalization of Chain of Thought, enables a new framework for deliberate decision-making by considering multiple different reasoning paths and self-evaluating choices to decide the next course of action, as well as looking ahead or backtracking when necessary to make global choices. For knowledge-intensive reasoning tasks, Knowledge-Augmented Reasoning Distillation \parencite{kang2023knowledgeaugmented} fine-tunes small LMs to generate rationales obtained from LLMs with augmented knowledge retrieved from an external knowledge base. Through this approach, significantly smaller models, such as 250M-parameter T5, can outperform fine-tuned large models with 12 times as many parameters.

PlanBench \parencite{valmeekam2022planbench} is an extensible benchmark suite maintained by the automated planning research community, covering diverse use cases including robotics, aerial-vehicle navigation, control systems, energy management and integrated-circuit design. The benchmark facilitates testing the capabilities of LLMs in planning actions and reasoning about dynamic changes. Currently, state-of-the-art LLMs are still struggling to be competitive on PlanBench.

\section{Methodology}
\label{methodology}
In this section, we first present our proposed framework for enhancing Q\&A AI performance. Based on the starting point of a typical RAG workflow, we identify key areas that could be improved. Following this identification, we briefly discuss the primary choices and key technical considerations for each area. We then propose a systematic design space map that can help AI teams be more deliberate and explicit in enumerating and making their technical choices when building Q\&A AI systems.

Next, we describe the \texttt{FinanceBench} dataset we used to develop and benchmark models and techniques. We also provide an overview of our evaluation methodology, which assesses the quality of outputs across different technical configurations to ensure robustness and effectiveness.

\subsection{Proposed Framework and Technical Design Space}
\label{framework-design-space}

A typical baseline RAG workflow today comprises the following steps:

\begin{enumerate}
    \item \textbf{Information Indexing and Retrieval} (the \textbf{``R''} step): The information sources are first indexed, typically into vector embeddings in the most common case of text-heavy sources. The typical means for indexing is an embedding model that encodes chunks of content into fixed-sized latent vectors that capture the semantic essence of such content. The same embedding model is then also used to search for content that is semantically relevant to a particular question. This retrieval step is typically based on similarity metrics or semantic matching algorithms.
    \item \textbf{Information Augmentation} (the \textbf{``A''} step): Either or both of the indexing and retrieving steps can be augmented with additional relevant metadata or human-provided considerations. Metadata can come from the formats and structures of the sources (e.g., PDF, CSV, etc.), while human-provided considerations may come from expert knowledge of notable facts, rules and exceptions, and process heuristics in the concerned domain. In certain domains with very specialized jargon, it is also desirable to augment the workflow to boost the attention to certain crucial keywords. The information augmentation would then influence the relevance rankings of the retrieved candidate content chunks, yielding more helpful supporting information for the downstream (Generation) step. The overall effect is to give firmer grounding for that subsequent Generation step, by helping index the information more appropriately, retrieve information more relevantly, and add external knowledge that the generative LLM does not natively have.
    \item \textbf{Answer Generation} (the \textbf{``G''} step): An LLM synthesizes a coherent and understandable answer using a combination of the retrieved supporting information and its own knowledge and generative capabilities. This method typically yields more accurate and less error-prone results than using the LLM alone.
\end{enumerate}

Such RAG workflows perform inferencing in one pass-through of these steps, from retrieving information relevant to the posed question, to augmenting such information with metadata and user-provided considerations, to generating an answer to the user.

We have identified, and shall elaborate in this paper, that there are potential gains in accuracy from the following changes to the RAG workflows:

\begin{enumerate}
    \item \textbf{A better Retriever}: an embedding model that more precisely encodes the latent meanings and connotations of certain important terms in domain-specific technical jargon;
    \item \textbf{A better Generator}: an LLM that “knows” the logics and desired presentational formats for the specific domain; and
    \item \textbf{Iterative, Multi-step Reasoning} employing RAG as an informational resource: a mechanism that combines RAG-based queries with necessity-and-sufficiency reasoning, task decomposition, and result verification, which are activities typically performed by humans to ensure accurate outcomes.
\end{enumerate}

Potential gains from other areas, such as from domain-specific Augmenters and from higher-level Hierarchical Task Planning (HTP) mechanisms, will be addressed in a future publication (see section \ref{conclusion}).

\subsubsection{Embedding Models for Indexing \& Retrieval}
\label{embedders}

\textit{Generic Embedding Models}

OpenAI's proprietary \texttt{text-embedding-ada-002}, released in December 2022, became the state of the art for information retrieval tasks such as text search, code search, and sentence similarity.. This embedding model offers unification of capabilities (merging five separate models into a simplified one) and longer context length (from 2048 to 8192). So far, OpenAI has not made available any means of fine-tuning \texttt{text-embedding-ada-002}.

BAAI's \texttt{bge-large-en}, an open-source embedding model, achieved first place on the Massive Text Embedding Benchmark leaderboard upon its release in August 2023 \parencite{zhang2023retrieve}. BAAI provides fine-tuning capabilities on this embedding model, which the LlamaIndex library provides an interface for. Its strong performance on general text-retrieval tasks makes it an ideal candidate for a baseline embedding model to observe the performance gain achieved by fine-tuning.

\textit{Fine-Tuned Embedding Models}

Fine-tuning an embedding model alters the mapping of text into a latent vector space. This text-to-vector mapping facilitates retrieval by identifying text chunks closest to the query within that space.

Within a given domain, certain terms may map to different meanings than they do in the colloquial sense, so fine-tuning the embedding model can change how such terms are associated given the context. For example, in a general embedding model, the word ``pizza'' might map closely to ``Italy,'' ``pasta,'' and other food-related concepts. However, in the context of winter sports, a fine-tuned model would associate it more with terms like ``slowing,'' ``stopping,'' and beginner-level skiing instructions.

In our development and experimentation, we have fine-tuned embedding models from BAAI’s \texttt{bge-large-\\en} model. The process generally involves splitting textual documents via the LlamaIndex library’s \texttt{Sentence\\Splitter} to create semantic chunks, and then using OpenAI’s GPT-3.5 to generate question-context pairs for each chunk for the synthetic dataset over which to fine-tune, and then running fine-tune training for 5 epochs. In use cases where there is already a natural question-and-resolution structure, such as in CRM systems, we can directly extract question-context pairs without needing to generate them using an LLM. In the particular use case of financial analysis covered in this paper, we select a random subset of 100 query-context pairs from the \texttt{FinanceBench} public dataset for the fine-tuning.

\subsubsection{Generative Models for Answer Generation}
\label{generators}

\textit{Generic LLMs}

OpenAI's proprietary GPT-3.5, launched in 2022, was a breakthrough in public adoption of LLM technology. This model has shown remarkable performance across various NLP tasks, serving as a robust foundation for Q\&A AI systems. Since then, a fast-growing number of LLMs, both commercial and open-source, have been released, creating a blooming ecosystem of LLM technology with increasingly diverse capabilities.

OpenAI's GPT-4, the company's most advanced LLM to date, accepts multi-modal input data including both text and images. Google DeepMind's proprietary Gemini models, offering many comparable capabilities, are direct competitors to OpenAI's GPT models. Startup Anthropic's Claude models have been widely adopted for their ability to handle very long contexts.

On the open-source front, Meta's Llama2 model family has seen the widest adoption, spawning an ecosystem of AI tools based around it. HuggingFace hosts a number of alternatives to Llama2, with the Falcon model family being among the most notable.

\textit{Fine-Tuned LLMs}

Fine-tuning an LLM requires optimal examples of conversational history including query-answer pairs. At inference time in a RAG system, the user’s message is augmented with document texts that the LLM can draw on to answer. However, fine-tuning an LLM typically focuses solely on query-answer pairs to isolate the answering behavior to depend only on the query, rather than the retrieved documents.

In our development and experimentation, we fine-tune on \texttt{gpt-3.5-turbo-0125} using OpenAI’s fine-tuning API.  In the particular use case of financial analysis covered in this paper, we select a random subset of 100 query-context pairs directly from the \texttt{FinanceBench} public dataset, then run fine-tune training for 5 epochs. For consistency we use the same random subset as is selected in the embedding model fine-tuning discussed above.

\newpage
\subsubsection{Iterative Reasoning}
\label{iteration}

\textit{OODA Reasoning Paradigm}

The Observe-Orient-Decide-Act (OODA) loop is a well-established iterative reasoning framework emphasizing continuous adaptation and decision-making in complex environments. Originally developed by military strategist John Boyd \parencite{boyd86patterns}, the OODA loop has been widely applied in various domains, including business, sports, and healthcare \parencite{enck2012ooda}.

The OODA loop consists of four main stages:

\begin{enumerate}
    \item \textbf{Observe}: Gather information about the environment and the problem at hand;
    \item \textbf{Orient}: Analyze the collected information, update the understanding of the situation, and generate potential solutions or actions;
    \item \textbf{Decide}: Evaluate the potential solutions or actions and select the most appropriate one based on the current understanding; and
    \item \textbf{Act}: Execute the selected solution or action and monitor its impact on the environment.
    \end{enumerate}
    
The iterative nature of the OODA loop allows for continuous refinement and adaptation based on the feedback received from the environment. By repeatedly cycling through these four stages, an agent can progressively improve its understanding of the problem, generate more relevant solutions, and make better decisions.

In the context of Q\&A systems, the OODA loop can be applied to decompose a hard question into simpler ones more easily satisfied by RAG queries, to combine and make sense of data points from multiple RAG queries, and to verify internal consistency. By incorporating the OODA loop, the Q\&A system can iteratively refine its understanding of the question, retrieve more relevant information, and generate more accurate and contextually appropriate answers. This iterative process helps overcome the limitations of one-pass input-process-output data flow, which characterizes many current LLMs and AI systems.

The integration of the OODA loop into RAG-based Q\&A systems involves mapping the four stages of the loop to the various components of the RAG pipeline. In the next section, we describe the construction of the OODA mechanism within our proposed framework.

\begin{figure}[h]
    \centering
    \includegraphics[width=0.65\textwidth]{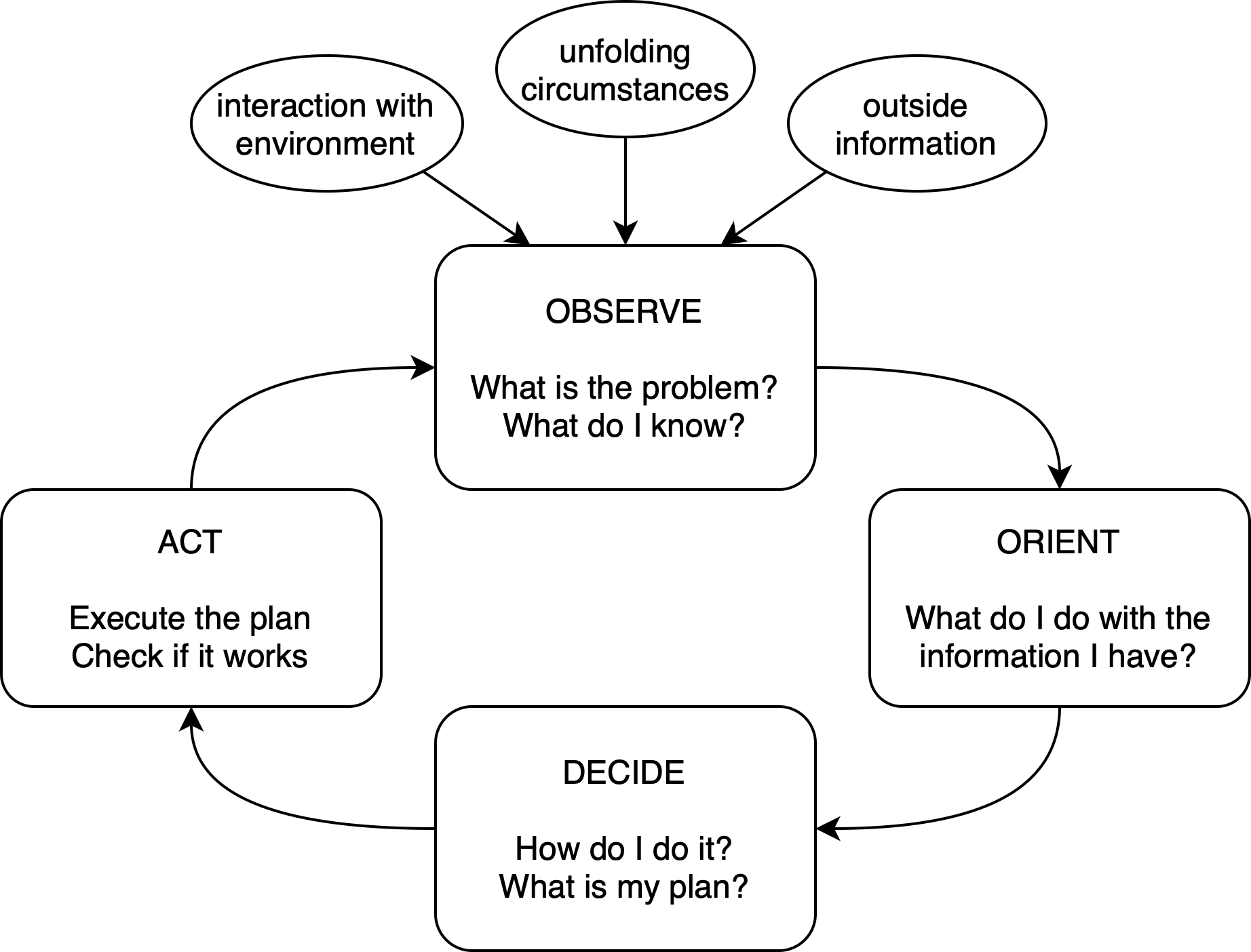}
    \caption{A typical OODA reasoning loop.}
    \label{fig:ooda}
\end{figure}

\textit{Constructing OODA Mechanism}

To clearly guide a generative model to reason through a task, we use the architecture implemented in Aitomatic’s open-source OpenSSA\footnote{\url{https://github.com/aitomatic/openssa}} framework, which incorporates OODA iterations. To execute on a task, a reasoner reasons through it and returns a conclusion. During the reasoning step, the model iteratively applies the OODA loop to the task at hand given the instructions and tools provided in the task description. This process involves continuously gathering data using the available resources (observe), reflecting on the current context and how the new information fits within it (orient), making decisions based on the accumulated knowledge and the goals of the task (decide), and then implementing those decisions using available tools (act). Iterating through OODA provides the model a consistent, clear framework for achieving task objectives, and allows it to adapt its approach as needed based on the available information.

\begin{figure}[h]
    \centering
    \includegraphics[width=\textwidth]{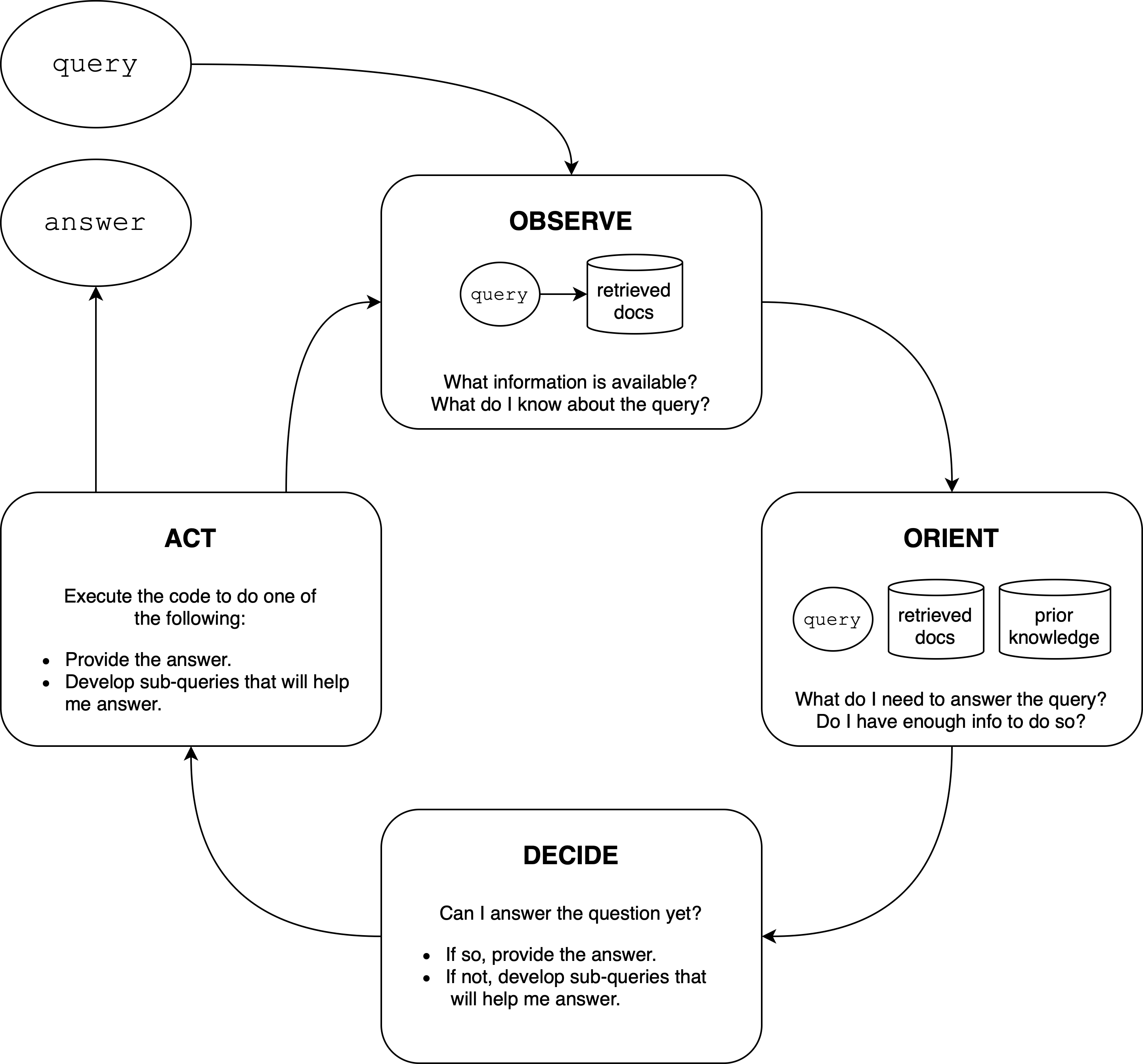}
    \caption{A specific implementation of OODA applied to question-answering with RAG.}
    \label{fig:ooda-rag}
\end{figure}

\begin{figure}[h]
    \centering
    \includegraphics[width=0.8\textwidth]{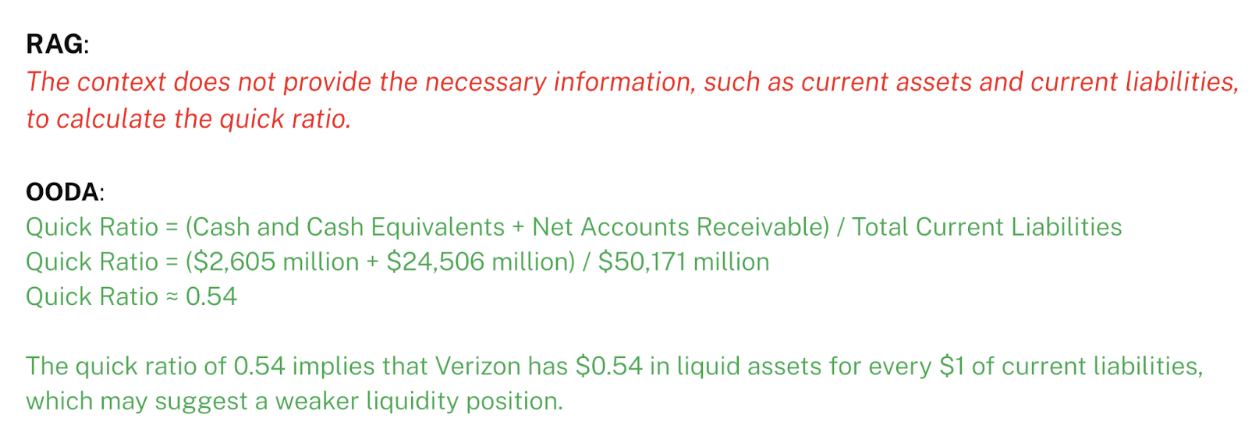}
    \caption{Comparison of pure-RAG and OODA-enabled answers to a \texttt{FinanceBench} question.}
    \label{fig:ooda-financebench}
\end{figure}

\newpage
\subsubsection{The Q\&A AI Technical Design Space}
\label{design-space}

Based on the technical considerations discussed above, we propose that AI developers and managers explicitly analyze a structured technical design space that outlines the principal components that meaningfully impact their AI workflows’ accuracy.

\begin{figure}[h]
    \centering
    \includegraphics[width=0.8\textwidth]{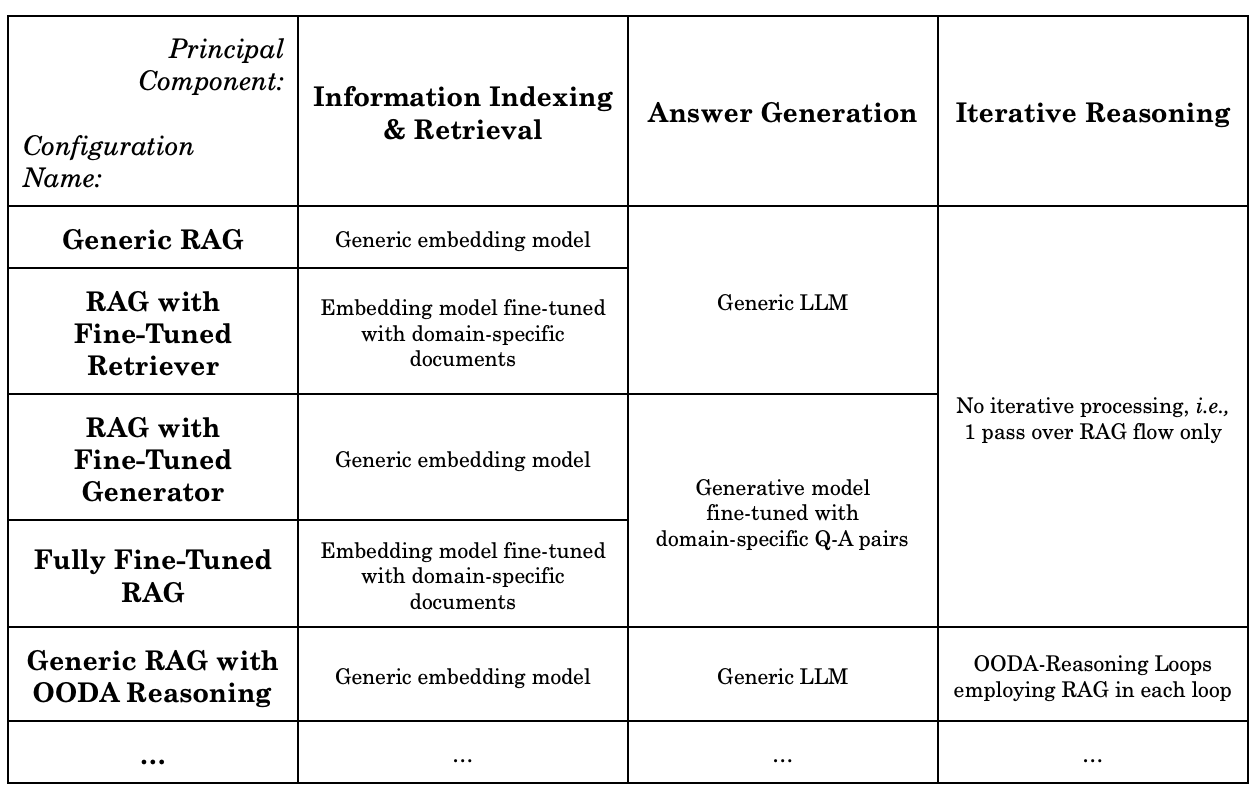}
    \caption{A structured technical design space capturing high-impact components within question-answering systems.}
    \label{fig:design-space}
\end{figure}

If we take into account just the three areas we have identified above for potential improvement, an example design space materializes as shown in Figure \ref{fig:design-space}.

As more principal technical components — such as Information Augmentation and Task Planning, which will be the subjects of our future publications — are identified as impactful, the design space should expand to include these as additional dimensions.

In the remaining sections of this paper, we explore specific configurations within the design space through experimental setups and present results that quantify the impacts of these technical choices. This exploration aims to provide AI developers and managers with concrete evidence to guide their decisions and enhance project success.

\subsection{Financial Analysis Benchmark Dataset}
\label{dataset}

\texttt{FinanceBench} is a financial-analysis dataset curated and partially open-sourced by \cite{islam2023financebench}. It sources its data from public company filings with the U.S. Securities and Exchange Commission (SEC), which typically include key financial statements (Balance Sheet, Income Statement, and Cash Flows Statement) and management commentaries on business outlook and risks. For each filing, one or several analytical questions are crafted, with expert answers provided by chartered financial analysts (CFAs).

\begin{figure}[h]
    \centering
    \includegraphics[width=\textwidth]{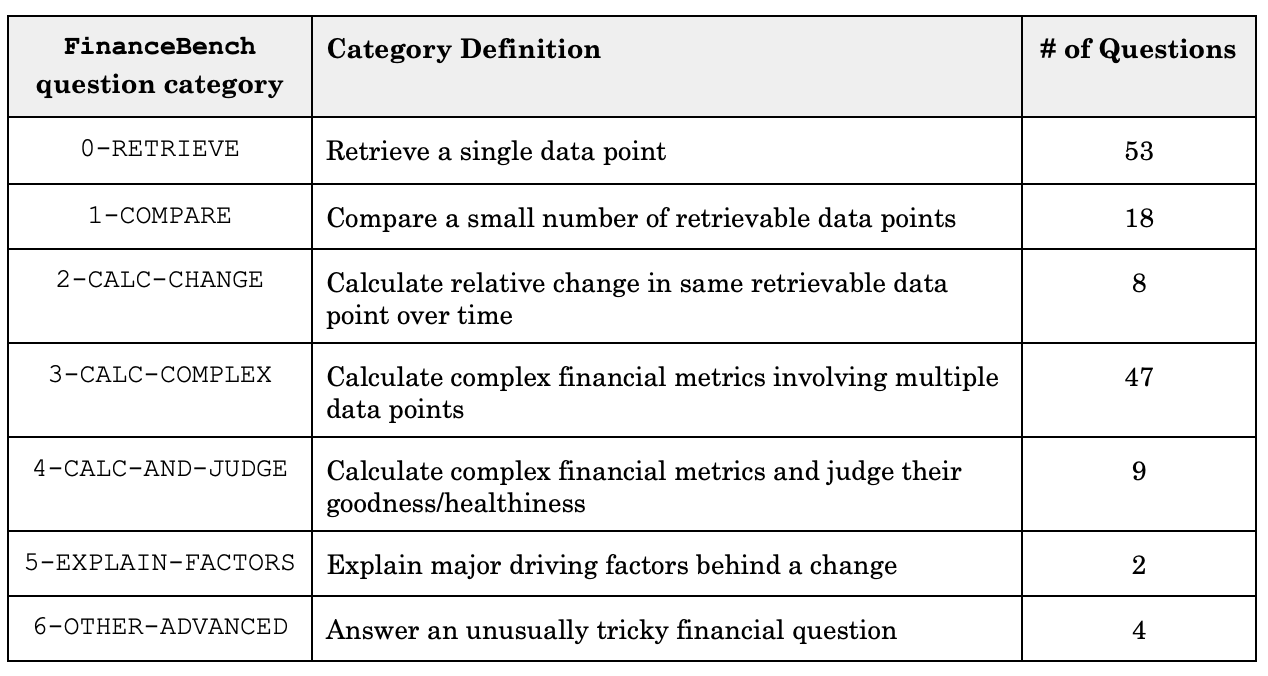}
    \caption{Question difficulty categorizations for \texttt{FinanceBench}.}
    \label{fig:financebench-categories}
\end{figure}

Out of 150 \texttt{FinanceBench} question-answer pairs made publicly available, 141 are suitable for benchmarking exercises (the remaining 9 are unusable due to broken/missing document sources). We have categorized these questions into seven categories of increasing difficulty, as illustrated in Figure \ref{fig:financebench-categories}.

\subsection{Evaluation Metrics}
To assess the quality of the outputs from our various technical combinations, we employ a range of evaluation metrics that consider both the accuracy and relevance of the generated responses.

\subsubsection{Retrieval Quality Metrics}
We use automated retrieval quality metrics from LlamaIndex’s evaluation suite, which employs an LLM to validate the retrieved documents as they relate to the query, response, and reference documents.
\begin{itemize}
    \item \textbf{Relevancy}. Relevancy measures whether or not the response for the query is in line with the context information. A response scores 0 if it is irrelevant and 1 otherwise.
    \item \textbf{Faithfulness}. Faithfulness measures whether a response is supported by the retrieved context, scoring 0 when it is not and 1 when it is.
    \item \textbf{Context similarity}. Context similarity measures the semantic difference between the retrieved contexts and the reference contexts. It is calculated by finding the cosine similarity between the contexts when mapped into an embedding space. By default, LlamaIndex rounds this score to 0 or 1 if the cosine is below or above 0.8 respectively.
\end{itemize}

\subsubsection{Answer Correctness Metrics}
We employ both an automated correctness metric and manual grading to evaluate whether the final answers produced by the Q\&A system given the reference answer.

\begin{itemize}
    \item \textbf{Automated correctness}. We use LlamaIndex’s \texttt{CorrectnessEvaluator}, which scores answers on a 1-5 scale based on their relevance to the reference answer and provides justifications for each score to ensure consistency. 1 signifies an irrelevant answer, 2-3 indicate a relevant answer with possible errors, and 4-5 denote a fully correct answer. For consistency with the manually-calculated binary scores, we convert these scores to percentages by subtracting 1 and dividing by 4.
    \item \textbf{Human correctness}. We also conduct human evaluations to assess the quality of the generated responses, rating answers as correct or incorrect based on the \texttt{FinanceBench} reference. To observe performance gains in different types of question-answering, we look at the performance in easier or harder question categories as described in Section \ref{dataset}. This qualitative assessment provides valuable insights into the real-world applicability and effectiveness of our proposed framework.
\end{itemize}

By employing these evaluation metrics, we aim to provide a comprehensive and rigorous assessment of the performance of our models in handling domain-specific Q\&A tasks. The combination of automated metrics and human evaluations allows us to thoroughly analyze the strengths and limitations of our proposed framework and identify areas for further improvement.

\section{Experiments \& Results}
\label{results}
We experimented with and evaluated the outputs of the following technical configurations:
\begin{itemize}
    \item Generic RAG: \texttt{bge-large-en} with \texttt{gpt-3.5-turbo-0125}.
    \item RAG with Fine-Tuned Generator: \texttt{bge-large-en} with fine-tuned \texttt{gpt-3.5-turbo-0125}.
    \item RAG with Fine-Tuned Retriever: fine-tuned \texttt{bge-large-en} with \texttt{gpt-3.5-turbo-0125}.
    \item Fully Fine-Tuned RAG: fine-tuned \texttt{bge-large-en} with fine-tuned \texttt{gpt-3.5-turbo-0125}.
    \item Generic RAG with OODA Reasoning: \texttt{bge-large-en} with \texttt{gpt-3.5-turbo-0125} and OODA reasoning.
\end{itemize}
Each system uses the same VectorStoreIndex for the documents, a top-k of 10 documents, and default RAG prompts, all from base LlamaIndex. For the full reproducible experimental setup, see OpenSSA's FinanceBench example\footnote{\url{https://github.com/aitomatic/openssa/tree/main/examples/FinanceBench}}. Other combinations such as Fine-Tuned RAG with OODA Reasoning shall be addressed in a future publication (see Section \ref{conclusion}).

Our test set consists of the remaining 41 questions in the publicly-available \texttt{FinanceBench} test of valid questions that were not the 100 selected for training. 

\subsection{Retrieval Quality Results}
Table \ref{tab:results-retrieval-41} summarizes the retrieval quality scores achieved by various configurations across the \texttt{FinanceBench} test set questions. As the OODA reasoning involves multiple iterative retrievals, it is not directly comparable to the one-step retrieval processes of these RAG systems, and is therefore not included in this table.

\begin{table}[h]
    \centering
    {\ttfamily \bgroup \def\arraystretch{1.5}
        \begin{tabular}{l|c|c|c}
             Configuration & Relevancy & Faithfulness & \makecell{Context\\similarity} \\ \hline
             Generic RAG & 0.317 & 0.700 & 0.894 \\
             Fine-Tuned Generator & 0.291 & 0.625 & 0.890 \\
             Fine-Tuned Retriever & 0.340 & 0.634 & 0.901 \\
             Fully Fine-Tuned RAG & 0.317 & 0.512 & 0.903 \\
        \end{tabular}
    \egroup}
    \caption{Retrieval quality metrics of question-answering systems on the \texttt{FinanceBench} test set.}
    \label{tab:results-retrieval-41}
\end{table}

\subsection{Answer Correctness Results}
Table \ref{tab:results-answer-41} summarizes the automated scores and human-judged binary-accuracy scores achieved by various configurations across the \texttt{FinanceBench} test set questions. Note that the automated correctness scores have been converted from their original 1-5 scaling to percentages. The \texttt{OVERALL} scores are performed on the entire test set. The scores for \texttt{EASIER} and \texttt{HARDER} questions are based on the respective subsets as defined in Section \ref{dataset}.

\begin{table}[h]
    \centering
    {\ttfamily \bgroup \def\arraystretch{1.5}
        \makebox[\linewidth][c]{
            \begin{tabular}{l|c|c|c|c}
                Configuration
                & \makecell{Human Evaluation\\\underline{EASIER QUESTIONS}\\0-RETRIEVE\\1-COMPARE\\2-CALC-CHANGE}
                & \makecell{Human Evaluation\\\underline{HARDER QUESTIONS}\\3-CALC-COMPLEX\\4-CALC-AND-JUDGE\\5-EXPLAIN-FACTORS\\6-OTHER-ADVANCED}
                & \makecell{Human\\Evaluation\\\underline{OVERALL}}
                & \makecell{Automated\\Correctness\\\underline{OVERALL}
                }
                \\ \hline
                Generic RAG & 55\% & 16\% & 37\%& 44\%\\ 
                Fine-Tuned Generator & 64\% & 16\% & 43\%& 45\%\\
                Fine-Tuned Retriever & 73\% & 35\% & 54\%& 54\%\\
                Fully Fine-Tuned RAG & 77\% & 37\% & 59\%& 58\%\\
                \makecell[l]{Generic RAG with OODA\\Reasoning} & 91\% & 79\% & 85\%& 77\%\end{tabular}
        }
    \egroup}
    \caption{Answer correctness metrics of question-answering systems on the \texttt{FinanceBench} test set.}
    \label{tab:results-answer-41}
\end{table}

\section{Discussion \& Analysis}
\label{analysis}
In this section, we analyze the implications of our experimental results, discuss key findings, and provide recommendations to enhance Q\&A AI system performance through domain-specific fine-tuning and retrieval-augmented generation (RAG) techniques.

\subsection{Key Findings}
Our experimental results demonstrate several important findings:

\textit{Gains in accuracy from RAG with fine-tuning component models}

\begin{enumerate}
    \item RAG with fine-tuned retriever, fine-tuned generator, or full fine-tuning outperforms the generic RAG. Specifically, using the \texttt{FinanceBench} dataset, we achieved accuracy improvements of up to 20 percentage points over baseline RAG.
    \item Fine-tuning the retriever model results in higher accuracy gains compared to fine-tuned generators. This advantage is significant as fine-tuning embedding models for retrieval is less costly and less labor-intensive than fine-tuning LLMs for generation.
\end{enumerate}

\textit{Gains in accuracy from RAG with OODA Reasoning}

\begin{enumerate}
    \item Integrating iterative reasoning capabilities, such as OODA, with the RAG engine substantially enhances performance. Specifically, the Generic RAG with OODA Reasoning configuration achieves an accuracy increase of \textbf{up to 50 percentage points} across the \texttt{FinanceBench} dataset compared to the generic RAG baseline.
    \item This finding underscores that OODA, though a domain-agnostic reasoning mechanism, significantly boosts accuracy in domain-specific tasks when integrated with relevant informational sources. On the \texttt{FinanceBench} dataset, the generic OODA configuration outperformed the fully fine-tuned RAG by a considerable margin of 20-25 percentage points.
\end{enumerate}

\subsubsection{Implications and Recommendations}
Our findings have significant implications for developing Q\&A AI systems across various domains. Based on our experimental results, we propose the following recommendations:

\begin{enumerate}
    \item Prioritize the fine-tuning of embedding models over language models. Our approach demonstrates superior performance and offers a more resource-efficient and scalable solution for addressing real-world challenges by capitalizing on domain-specific embedding tuning. Fine-tuning embedding models requires minimal human effort for data acquisition compared to fine-tuning language models, making it a more practical choice.
    \item Employ OODA reasoning or other iterative reasoning mechanisms which can significantly improve informational consistency and enhance the Q\&A system's ability to combine information from multiple sources. 
    \item Map out a technical design space covering principal components and make deliberate choices based on empirical evidence: (a) Identify the principal components that have the most significant impact on the performance of Q\&A systems, such as information indexing and retrieval, answer generation, and iterative reasoning. (b) Create a structured technical design space that captures the possible configurations of these components and (c) Make informed decisions on the optimal configuration for a given domain or task based on empirical quantitative evidence, such as the results reported in this study.
\end{enumerate}

By following these recommendations, AI teams can leverage the potential of domain-specific fine-tuning and iterative reasoning to enhance the accuracy, relevance, and usefulness of generated responses, while maintaining resource efficiency and scalability.

\section{Conclusions \& Future Directions}
\label{conclusion}

In this paper, we identified areas where new or modified techniques can significantly improve the accuracy of LLM-based RAG in Q\&A AI use cases. In particular, we have investigated the how-tos and the impacts of domain-specific fine-tuning and iterative reasoning, measuring accuracy gains from such methods on complex financial-analysis and industrial-equipment datasets. Promising accuracy improvements of up to 30 percentage points on the generic RAG baseline suggest there is a lot of value in incorporating these methods in real-world applications in specialized domains.

Additionally, we have elaborated that designing a high-quality AI system involves many important technical choices, and that AI developers and managers can benefit from a structured technical design space map that helps them be very deliberate and informed in their decision-making.

Looking ahead, we identify several promising directions for future research and development:

\begin{enumerate}
    \item \textbf{Custom Augmenters for RAG:} Explore the development of domain-specific augmenters that leverage domain-specific metadata and expert knowledge such as notable facts, rules and exceptions, and process heuristics, to prioritize and filter the most pertinent information for a given question.
    \item \textbf{Fine-Tuned RAG with OODA Reasoning}: Quantify the accuracy gains from combining fully fine-tuned RAG with iterative OODA loops.
    \item \textbf{Practical Guideline for AI Teams}: Develop a set of actionable practices for AI teams seeking to build high-performance Q\&A systems. This guideline will be based on the growing data catalog contributed by the AI Alliance members, the technical design space mapping framework, and quantitative empirical results measuring performance of configurations in that design space on datasets from the catalog. Through the AI Alliance, we all can ensure open development, open science, and open AI, benefiting the broader AI community.
    \item \textbf{Advanced Planning \& Reasoning}: With our current experimentation showing great promise from integrating OODA reasoning, we are continuing our research to explore Hierarchical Task Planning (HTP) and other planning mechanisms that are well known in financial and industrial domains. Well-performing and efficient implementations of Planning and Reasoning would push the boundaries of problem-solving capabilities in mission-critical real-world AI systems.
    \item \textbf{Open-Source Industry-Specific Evaluation Datasets}: With the proliferation of applications of AI to various industries, it would be highly beneficial to establish meaningfully complex benchmark datasets for key industries or specific domains, especially those with scientific or engineering knowledge and technical jargon and that significantly differ from general knowledge and language.
\end{enumerate}

\section{Acknowledgements}
We would like to express our gratitude to the AI Alliance (\url{https://thealliance.ai}) for providing the impetus and platform for collaboration in open science. We also extend our thanks to the member organizations of the AI Alliance, their researchers and engineers for their valuable contributions to this study. Their expertise, insights, and collaborative spirit have been instrumental in advancing our research.

\printbibliography

\section*{Appendix}

\subsection*{Full set results}

We also conducted tests on the full publicly-available FinanceBench valid question set, which includes questions that were used in fine-tuning training, and found consistent results with the test set results for answer quality. Results are shown in table \ref{tab:results-retrieval-full} and \ref{tab:results-answer-full}.

\begin{table}[h]
    \centering
    {\ttfamily \bgroup \def\arraystretch{1.5}
        \begin{tabular}{l|c|c|c}
             Configuration & Relevancy & Faithfulness & \makecell{Context\\similarity} \\ \hline
             Generic RAG & 0.354 & 0.709 & 0.887 \\
             Fine-Tuned Generator & 0.355 & 0.673 & 0.887 \\
             Fine-Tuned Retriever & 0.426 & 0.738 & 0.901 \\
             Fully Fine-Tuned RAG & 0.418 & 0.723 & 0.901 \\
        \end{tabular}
    \egroup}
    \caption{Retrieval quality metrics of question-answering systems on the \texttt{FinanceBench} full set.}
    \label{tab:results-retrieval-full}
\end{table}

\begin{table}[h]
    \centering
    {\ttfamily \bgroup \def\arraystretch{1.5}
        \makebox[\linewidth][c]{
        \begin{tabular}{l|c|c|c|c}
             Configuration
             & \makecell{Automated\\correctness\\score\\(1-5)} 
             & \makecell{Human evaluation\\\underline{EASIER QUESTIONS}\\0-RETRIEVE\\1-COMPARE\\2-CALC-CHANGE}
             & \makecell{Human evaluation\\\underline{HARDER QUESTIONS}\\3-CALC-COMPLEX\\4-CALC-AND-JUDGE\\5-EXPLAIN-FACTORS\\6-OTHER-ADVANCED}
             & \makecell{Human\\evaluation\\\underline{OVERALL}} 
             \\ \hline
             Generic RAG & 2.83 & 63\% & 18\% & 43\% \\ 
             Fine-Tuned Generator & 2.91 & 65\% & 20\% & 45\% \\
             Fine-Tuned Retriever & 3.23 & 70\% & 35\% & 55\% \\
             Fully Fine-Tuned RAG & 3.21 & 68\% & 37\% & 55\% \\
             \makecell[l]{Generic RAG with OODA\\Reasoning} & 3.91 & 91\% & 69\% & 82\%
        \end{tabular}}
    \egroup}
    \caption{Answer correctness metrics of question-answering systems on the \texttt{FinanceBench} full set.}
    \label{tab:results-answer-full}
\end{table}

\subsection*{Generated training data}

We experimented with fine-tuning on synthetic training data. We first collected the PDFs from all links listed in the \texttt{FinanceBench} dataset and split their texts into chunks of 1024 tokens each with LlamaIndex’s SentenceSplitter. Then, we used LlamaIndex’s RagDatasetGenerator to generate 3 queries and answers per chunk. In total, this created about 15,500 query-context-answer triplets that could then be used for fine-tuning embeddings (query-context) and fine-tuning generators (query-answer). For smaller fine-tuning jobs we used random subsets of this dataset to train.

We found that performance either did not improve or decreased when we used models fine-tuned on this data. The likeliest explanation, given the success of fine-tuning on data taken directly from \texttt{FinanceBench}, is that when GPT-3.5 generates queries and answers based on the contexts, it does not do so in the way that the domain experts that created the \texttt{FinanceBench} data did. These may lead to questions that rely more heavily on the wording in the context, answers that do not use prior financial knowledge in addition to the contexts, or other quirks of generated data that do not correspond to the ways humans think about the relationships between texts for synthesis tasks. The results are shown in Table \ref{tab:results-synth}

\begin{table}[h]
    \centering
    {\ttfamily \bgroup \def\arraystretch{1.5}
        \makebox[\linewidth][c]{
            \begin{tabular}{ll|c}
                 \makecell[l]{Embedder\\(+ train size)} & 
                 \makecell[l]{Generator\\(+ train size)} & 
                 \makecell{Automated\\correctness}
                 \\ \hline
                 Generic & Generic & 2.83 \\ \hline
                 Generic & Fine-tuned on 16k & 2.57 \\
                 Fine-tuned on 16k & Generic & 2.44 \\
                 Fine-tuned on 16k & Fine-tuned on 16k & 2.34 \\ \hline
                 Generic & Fine-tuned on 1k & 2.95 \\
                 Fine-tuned on 1k & Generic & 2.96 \\
                 Fine-tuned on 1k & Fine-tuned on 1k & 2.96 \\
            \end{tabular}
            }
    \egroup}
    \caption{Correctness scores of fine-tuning configurations on the \texttt{FinanceBench} test set.}
    \label{tab:results-synth}
\end{table}

\end{document}